\documentclass[10pt,twocolumn,letterpaper]{article}

\usepackage[pagenumbers]{cvpr} % To force page numbers, e.g. for an arXiv version

\definecolor{cvprblue}{rgb}{0.21,0.49,0.74}
\usepackage[pagebackref,breaklinks,colorlinks,allcolors=cvprblue]{hyperref}

\def\paperID{15182} % *** Enter the Paper ID here
\def\confName{CVPR}
\def\confYear{2025}

\title{VIST-GPT: Ushering in the Era of Visual Storytelling with LLMs?}

\author{Mohamed Gado\hspace{0.7cm}Towhid Taliee\hspace{0.7cm}Muhammad Memon\hspace{0.7cm}Dmitry Ignatov\hspace{0.7cm} Radu Timofte
\\\\
Computer Vision Lab, CAIDAS, University of Würzburg, Germany
}

\begin{document}
\maketitle
\begin{abstract}

Visual storytelling is an interdisciplinary field combining computer vision and natural language processing to generate cohesive narratives from sequences of images. This paper presents a novel approach that leverages recent advancements in multimodal models, specifically adapting transformer-based architectures and large multimodal models, for the visual storytelling task. Leveraging the large-scale Visual Storytelling (VIST) dataset, our VIST-GPT model produces visually grounded, contextually appropriate narratives. We address the limitations of traditional evaluation metrics, such as BLEU, METEOR, ROUGE, and CIDEr, which are not suitable for this task. Instead, we utilize RoViST and GROOVIST, novel reference-free metrics designed to assess visual storytelling, focusing on visual grounding, coherence, and non-redundancy. These metrics provide a more nuanced evaluation of narrative quality, aligning closely with human judgment.

% Instead, we utilize new reference-free metrics designed to assess visual storytelling, focusing on visual relevance. We also utilize a single metric to evaluate how closely our models' performance matches that of human-generated stories, achieving state-of-the-art results in this area.

\end{abstract}
\section{Introduction}

Visual storytelling is a compelling application of artificial intelligence that merges computer vision and natural language processing to create coherent narratives from image sequences. This emerging field captures and integrates visual and linguistic information to create engaging stories, enhancing our interaction with visual content and opening new possibilities in entertainment, education, and communication \cite{huang2016visual}. By mimicking how humans interpret visual content, visual storytelling transforms images into meaningful narratives  \cite{gonzalez2018contextualize, kim2018glac}.

A core challenge of visual storytelling lies in understanding both the content of images and the sequential relationships between them. Traditional computer vision techniques primarily focus on recognizing objects, actions, and scenes, but effective storytelling requires more than simple recognition. It necessitates the ability to craft temporally coherent and contextually relevant narratives that bridge the gap between literal descriptions and deeper storytelling  \cite{xu2021imagine, hong2020diverse}.

Additionally, visual storytelling excels in inferring emotions and deeper meanings from images, allowing for the transformation of basic descriptions into dynamic, human-like narratives \cite{liu2020character}. For example, interpreting an image of someone "sitting on a bench" as reflecting feelings of boredom enriches the storytelling experience, making it more engaging and relatable. This ability to convey emotional depth is essential for creating immersive narratives that resonate with audiences \cite{yang2022re3}.

%%Danish%%

% Visual storytelling takes a fresh approach about how we engaged with the images, allowing technology to craft genuine stories from a sequence of images. Unlike simple image captioning, which describes each picture, visual storytelling creates a flowing narrative, picking up on emotions, relationships, and the context around each scene \cite{lu2016visual}\cite{narayan2021planning}. The concept of transforming fields like entertainment, education, and digital media, turning images into something richer, deeper, and more connected to our human way of interpreting world.

% It’s not enough to simply see what’s in each image; it has to understand how these images connect, like chapters in a book. Traditional image analysis can only get so far by naming objects or identifying settings. But to weave a story, a system has to grasp the sequence that “what happened next” aspect that makes stories interesting \cite{rashkin2020plotmachines}. Visual storytelling demands a sense of flow and timing, drawing on both the obvious parts of each image and the more subtle cues in between \cite{puduppully2022data}.

Beyond just recognizing objects and actions, good storytelling also captures the mood and meaning behind them. For example, a picture of someone sitting alone on a bench in a quiet park might suggest feelings like loneliness, peace, or thoughtfulness. These kinds of details make the story feel richer and more relatable \cite{chen2021commonsense}. By capturing this sense of emotion, visual storytelling creates something that’s more than just a factual description that invites people to connect with the story in a personal way.

The impact of the visual storytelling can be seen in the different fields. In entertainment, it can be used to develop narratives for movies, video games, or for virtual experiences, allowing the creators to experiment with visual storytelling in a way that’s new and creative. In education, it could help students better connect with historical events or complex concepts through stories rather than static facts. For businesses and marketers, visual storytelling allows brands to communicate on an emotional level, turning their message into something that feels relatable and memorable \cite{shen2023storygpt}.

% Visual storytelling combines visuals and language in a complex but rewarding way. Early systems focused mostly on recognizing objects within images, while recent methods use more advanced models that capture the progression and depth needed for storytelling \cite{huang2016visual}. These stories are becoming better when it comes to recognizing emotions in the sequences, making stories they generate feel more natural and more engaging \cite{tian2024mm}.

Visual storytelling is opening the doors to a new way of interacting with the digital content one that goes beyond seeing images and starts to tell us the stories we can connect with. As the field develops, it’s exciting to think of where it could lead, shaping our interactions with technology in ways that feels closer to the way we, as humans, experience and remember stories.
\section{Dataset}

Visual Storytelling (VIST) dataset \cite{huang2016visual} was introduced to foster advancements in AI systems’ ability to create coherent, human-like stories based on a series of related images. Unlike traditional image captioning datasets, which focus on single image descriptions, VIST emphasizes the concept of storytelling across a sequence of images. It is structured specifically to capture the flow of events and varied connections that unfold over multiple images, providing a comprehensive benchmark for sequential narrative generation.

% Each story in the VIST dataset is consisting of a sequence of five images that depict real-life events or activities. These sequences are crafted to include everyday experiences, such as family gatherings, vacations, outdoor adventures, and other moments that naturally lend themselves to storytelling. The dataset’s design challenges models to not only identify objects and actions within each image but also to understand the temporal and contextual relationships across the sequence.

Each image sequence in VIST has a five-sentence narrative, one sentence per image. These sentences do more than just describe what is happening in each individual image; they weave together to create a coherent and engaging story that gives context to the events as a whole. These narrative approach reflects the way humans tend to recount experiences, linking each moment to build a larger story rather than just focusing on isolated descriptions.

% % The primary goal of the VIST dataset is to help the AI models in developing the ability to generate the narratives which focuses on the continuity and the coherence required to tell story. This capability is essential for tasks in visual storytelling, where AI is tasked with generating narratives from a set of images rather than simply captioning them.

% This dataset is particularly valuable in the development of models that need to understand. Recognizing how one image relates to the next in a sequence. Inferring the broader context that connects the images, such as a sequence of family members interacting at a party. Understanding and conveying the emotional undertones that a human storyteller might include, like joy, surprise, or introspection.

% VIST dataset has been instrumental in advancing visual storytelling research, it has some limitations. Each story consists of only five images, which limits the dataset versatility in representing longer or shorter narratives. The dataset’s focus on real-life events and activities means it may lack coverage for more abstract or imaginative storytelling scenarios. While VIST captures basic storytelling, its narratives may lack the depth and range of emotions needed for highly detailed storytelling.

\begin{figure*}[h]
    \centering
    \includegraphics[width=1.00\textwidth]{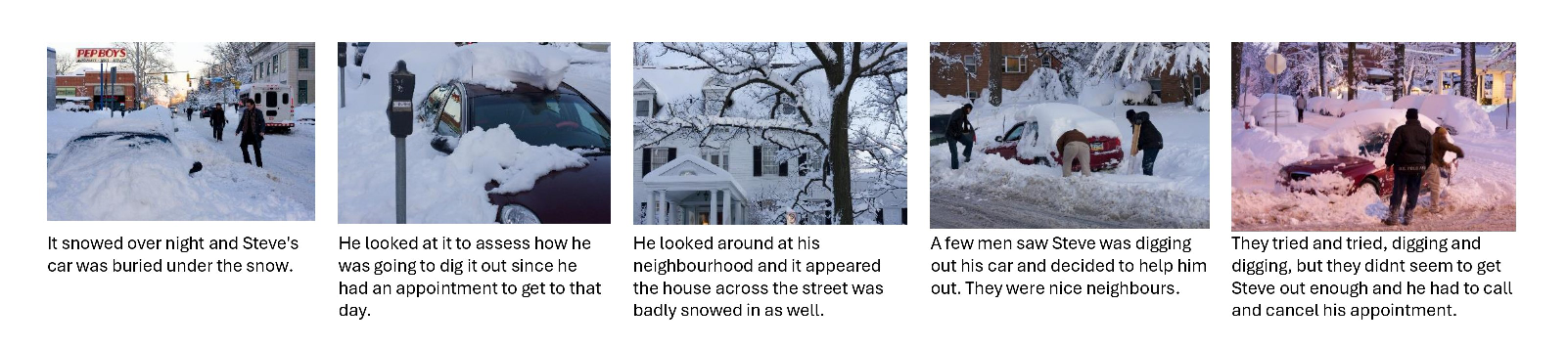}
    \caption{A sample from the VIST dataset consists of five images with their corresponding story sentences.}
    \label{fig:sample}
\end{figure*}

Visual Storytelling (VIST) dataset is very crucial in advancing AI’s storytelling capabilities, enabling systems to move beyond single image captions and generate cohesive, human-like stories from image sequences. An example from the dataset is shown in Figure \ref{fig:sample}.

\section{Related works}
\label{sec:related_works}

\subsection{Visual Storytelling}
% The field of visual storytelling has been developed as a means for AI to gain a human-like understanding of event structures and linguistic capabilities that are extend beyond mere description. 
Huang et al. (2016) \cite{huang2016visual} was among the first to introduce the Visual Storytelling (VIST) dataset, establishing a benchmark for generating coherent narratives from image sequences. The earlier models for visual storytelling, such as those by Gonzalez-Rico and Fuentes-Pineda (2018) \cite{gonzalez2018contextualize} and Kim et al. (2018) \cite{kim2018glac}, employed simple encoder-decoder architectures, with convolutional neural networks (CNNs) for feature extraction and recurrent neural networks (RNNs) for text generation available in the LEMUR NN Dataset \cite{ABrain.NN-Dataset}. Later research, including work by Xu et al. (2021) \cite{xu2021imagine}, Chen et al. (2021) \cite{chen2021commonsense}, and Hsu et al. (2020) \cite{hsu2020knowledge}, has incorporated external resources, such as ConceptNet, to embed Commonsense Knowledge into storytelling systems, enhancing their ability to generate narratives with realistic and intuitive reasoning. 

Alternative methods employ scene graphs to represent object relationships, enhancing the system's understanding of narrative structure (Lu et al., 2016 \cite{lu2016visual}; Hong et al., 2020) \cite{hong2020diverse}. A notable method by Hsu et al. (2021) \cite{hsu2021plot} involves building a graph from image sequences using training data and external resources to derive the best storyline path. Most of these visual storytelling models are trained from scratch (Xu et al., 2021 \cite{xu2021imagine}; Hsu et al., 2020 \cite{hsu2020knowledge}; Wang et al., 2020 \cite{wang2020storytelling}; Yang et al., 2019) \cite{yang2019knowledgeable}, though Chen et al. (2021) \cite{chen2021commonsense} experimented with a BART model as a baseline. By contrast, our approach leverages pretrained language models’ generalization capabilities to enhance visual storytelling.

\subsection{Planning and Story Generation}
The process of planning has become an effective approach to structuring story generation by organizing the narrative’s content. In automatic story generation, the process is often broken down into the two phases: creating an outline and then expanding it with details (Yao et al., 2019 \cite{yao2019plan}; Xu et al., 2018 \cite{xu2018skeleton}; Rashkin et al., 2020 \cite{rashkin2020plotmachines}). Plans have been represented through keywords (Yao et al., 2019) \cite{yao2019plan}, character actions (Liu et al., 2020) \cite{liu2020character}, and plot structures (Goldfarb-Tarrant et al., 2020a) \cite{goldfarb2020content}, with more recent methods incorporating details about the setting, characters, and plot points (Yang et al., 2022) \cite{yang2022re3}. The concept of an independent planning stage has been applied beyond storytelling to other text generation domains, such as summarization (Narayan et al., 2021) \cite{narayan2021planning} and data-to-text generation (Moryossef et al., 2019 \cite{moryossef2019step}; Puduppully et al., 2022 \cite{puduppully2022data}).

% approach integrates elements of story planning by encoding visual entities, organizing events temporally, and guiding the LLM through structured prompting and fine-tuning, allowing it to generate stories that align with the logical and narrative coherence sought in traditional storytelling frameworks.

\subsection{Multimodal Large Language Models}

The continuous advancement and application of large language models for various types of tasks \cite{ABrain.HPGPT, ABrain.NNGPT, Rupani2025llm} includes recent developments in multimodal large language models (MLLMs) are redefining the potential for integrating visual and linguistic understanding in AI. Early work, like Frozen (Tsimpoukelli et al., 2021) \cite{tsimpoukelli2021multimodal}, enabled zero-shot multimodal generation by combining gated cross-attention mechanisms with LLMs. Building on this, models like BLIP-2 (Li et al., 2023) \cite{li2023blip}, MiniGPT-4 (Zhu et al., 2023) \cite{zhu2023minigpt}, and LLaVA (Liu et al., 2023) \cite{liu2024visual} incorporated vision transformers (ViT) alongside LLaMA (Touvron et al., 2023) \cite{touvron2023llama} and Vicuna (Chiang et al., 2023) \cite{chiang2023vicuna} models to achieve single-image comprehension capabilities. NextGPT (Wu et al., 2023) \cite{wu2023next} introduced more flexible instruction-following abilities using any-to-any multimodal training data. Specifically for storytelling, MiniGPT-5 (Zheng et al., 2023) \cite{zheng2023minigpt} proposed using “vokens” (Tan and Bansal, 2020) \cite{tan2020vokenization} to allow text and image generation simultaneously, although challenges remain in maintaining consistent storytelling and narrative expressiveness. StoryLLaVA\cite{yang-etal-2025-storyllava}, a recent model specifically designed for visual storytelling, also suffers from hallucination, introducing objects, events, or relationships that are absent in the visual input. These hallucinations undermine narrative accuracy and coherence, making the generated stories less reliable. In contrast, VIST-GPT incorporates explicit narrative structure learning, fine-tunes an LLM for story generation, and applies rigorous evaluation metrics to ensure visual grounding and factual consistency. Our results show that fine-tuning significantly reduces hallucinations compared to both general-purpose VLMs and existing visual storytelling models, leading to more accurate and immersive storytelling.

\section{Methodology}

% Initially, we attempted to adapt LLaVA to the visual storytelling task through prompt
% engineering. However, LLaVA does not support multiple images as input. When we tried

% We used another model called VideoGPT+ \cite{maaz2024videogpt+}, which integrates both image and video encoders to enhance image understanding. Given that it's designed and trained for videos it would be good for image sequences in visual storytelling. We tried it with prompt engineering but it would generate more of a descriptions instead of naturally flown coherent stor

% We adapted the VideoGPT+ model \cite{maaz2024videogpt+} to the visual storytelling task, integrating both image and video encoders to enhance image understanding and finetuned it on VIST dataset.

Initially, we explored adapting LLaVA \cite{liu2024visual} to the visual storytelling task through prompt engineering. However, LLaVA does not support multiple image inputs. When we attempted to process five images simultaneously, the model treated them as a single composite image rather than interpreting them as distinct frames in a sequence.

To address this limitation, we experimented with VideoGPT+ \cite{maaz2024videogpt+}, a model that integrates both image and video encoders to enhance visual understanding. Since VideoGPT+ is designed for video processing, it is well-suited for handling sequential images in visual storytelling. However, our initial attempts using prompt engineering resulted in descriptive outputs rather than naturally flowing, coherent narratives.

To further refine the model for visual storytelling, we adapted VideoGPT+ by fine-tuning it on the VIST dataset, enhancing its ability to generate structured and contextually coherent narratives from image sequences.

\subsection{Architecture}

\begin{figure*}
    \centering
    \includegraphics[width=1.0\textwidth]{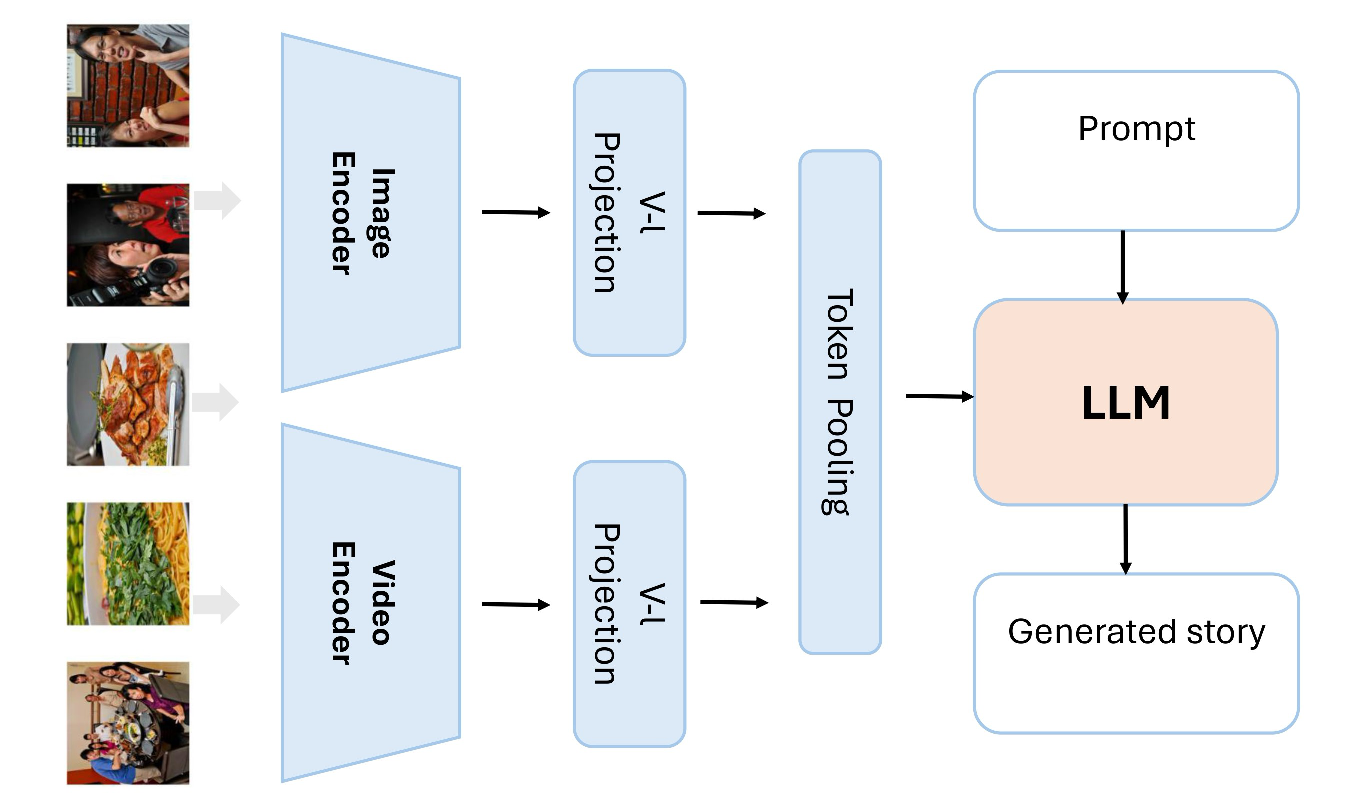}
    \caption{VIST-GPT Architecture (components in blue are frozen, while the LLM in red is fine-tuned with LoRA)}
    \label{fig:VST-GPT}
\end{figure*}

As shown in Figure \ref{fig:VST-GPT}, the model consists of:

\textbf{Dual Visual Encoder}: The model processes a sequence of images within a story, using an image encoder (CLIP ViT-L/14) \cite{radford2021learning} to capture detailed spatial features and a video encoder (InternVideo-v2) \cite{wang2024internvideo2} to model temporal dynamics. These two encoders are integrated to combine their strengths.

\textbf{Vision-Language (V-L) Adapter}: The features extracted from the image and video encoders are projected to a common space using V-L adapters (trainable MLPs). These adapters map the visual features to the language domain.

\textbf{Token Pooling}: Token pooling is applied to visual features to reduce the size without losing important information, allowing more data to fit into the LLM’s context without sacrificing crucial information.

\textbf{Large Language Model (LLM)}: The visual features (spatial and temporal) and user text input are connected and processed by the LLM (fine-tuned using LoRA). The LLM then generates responses based on the image and text inputs. We used the Phi-3-mini-4k-instruct \cite{abdin2024phi}, a compact, instruction-following language model designed for efficiency and fast performance, ideal for use in resource-limited environments.

VIST-GPT introduces several key innovations that enhance its performance in visual storytelling. The model employs a dual visual encoder architecture, integrating CLIP ViT-L/14 for spatial feature extraction and InternVideo-v2 for temporal dynamics, enabling a comprehensive understanding of both individual images and their sequential relationships. To achieve seamless multimodal integration, a Vision-Language Adapter is utilized to effectively align visual and linguistic embeddings, ensuring that generated narratives remain both visually grounded and contextually coherent.

To optimize model performance, Low-Rank Adaptation (LoRA) is employed for fine-tuning the Phi-3-mini-4k-instruct LLM, striking a balance between generalization and task-specific adaptation. Additionally, task-specific prompting enforces a one-sentence-per-image structure, with carefully tuned parameters—such as a reduced temperature setting and constrained beam search—to enhance narrative focus while minimizing redundancy.

\subsection{Training Strategy}

% For fine-tuning on the VIST dataset, we kept the model’s visual encoders (both image and video) frozen, along with the visual adapters, which were initially trained during the instruction-tuning stage with VideoGPT+. The large language model (LLM) was fine-tuned using a next-token prediction loss with LoRA on the VIST dataset.

The model’s visual encoders (image and video) remain frozen, while the visual adapters, initially trained during the instruction-tuning stage with VideoGPT+, are also kept frozen during fine-tuning on VIST. The LLM is fine-tuned on the VIST dataset using a next-token prediction loss with LoRA.

Each story was processed as a sequence of five images, treated as video frames. We used "InternVideo2-Stage2 1B-224p-f8" \cite{wang2024internvideo2}, which processes 8-frame sequences, padding the remaining three frames to fit the model’s input format. The image and video encodings were projected into the LLM space. The prompt used for generating stories was:

% For fine-tuning the model on the VIST dataset, we processed the data as a sequence of five images treated as video frames, utilizing "InternVideo2-Stage2 1B-224p-f8" \cite{wang2024internvideo2}, which processes 8 frames. The remaining 3 frames were padded. The prompt used for generating stories was:

\textbf{"video: Generate a short coherent story about the given sequence of images."}

The image and video encodings of the sequence were passed through the projection layers into the LLM space before the prompt. The model was trained for 5 epochs with a batch size of 4 -as each example already included 8 frames to process- on an NVIDIA DGX A100 (80 GB GPU memory), completing training in approximately 4 hours.

After completing the training process, we conducted several experiments during the inference phase, testing various parameters and prompts to optimize the model's performance. This experimentation resulted in two distinct versions of the model, VIST-GPT v1 and VIST-GPT v2. The prompts and parameters used for each version are shown in Table \ref{tab:videogpt_versions}.

Temperature controls randomness in text generation; higher values make responses more diverse, while lower values make them more focused. Number of Beams determines how many candidate outputs the model considers; more beams improve quality but increase computation, while fewer beams speed up inference with less variation. VIST-GPT v1 used a higher temperature and more beams, promoting more creative and varied storytelling. In contrast, VIST-GPT v2 used a lower temperature and fewer beams, leading to more structured and deterministic narratives. Additionally, VIST-GPT v2's refined prompt explicitly constrained the model to generate five sentences, each corresponding to a specific image, ensuring better coherence and alignment between the images and the generated story.

\begin{table}[htbp]
\centering
\setlength{\tabcolsep}{6pt} % Adjusts column separation
\renewcommand{\arraystretch}{1.2} % Adjusts row separation
\begin{tabular}{|l|p{2cm}|p{3cm}|} % Added vertical lines for better clarity
    \hline
    \textbf{Parameters} & \textbf{VIST-GPT v1} & \textbf{VIST-GPT v2} \\
    \hline
    Temperature & 0.8 & 0.7  \\
    \hline
    Number of Beams & 4 & 2  \\
    \hline
    Prompt & Generate a short coherent story about the given sequence of images & Given a sequence of five images, write a short coherent story of five sentences, with each sentence corresponding to an image.  \\
    \hline
\end{tabular}
\caption{Parameters used for VIST-GPT versions.}
\label{tab:videogpt_versions}
\end{table}

\begin{figure*}[h]
    \centering
    \includegraphics[width=1\textwidth]{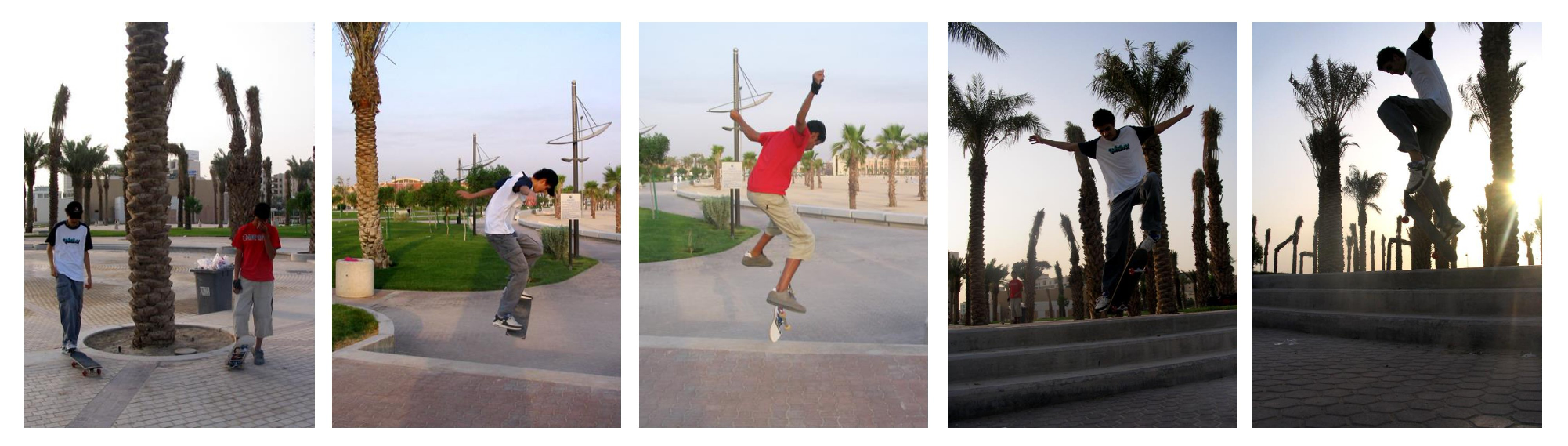}

\begin{tabular}{ l  p{13cm}} % Adjust the width (e.g., 4cm) as needed
    \hline
    Human Story & Hanging out with Abraham, getting some skate time in. Hit an olie of a two step, just warming up. Abraham hit this sweet kickflip! It made for a great photo. Manual up to a three step...into a frontside kickflip! My first time hitting this trick! I was super excited.  \\
    \hline
    AREL Model & I went to the park yesterday. The man is walking down the street to the park. The man in the middle of the race was a bit nervous, but he wasn't sure what he was doing. The man is trying to figure out what to do. This is a picture of a man. \\
    \hline
    GLACNET Model & The man was excited to be at the ski resort. He was very excited for his first day of skiing. He was able to make sure that he wasn't going fast. After a long ride, it was time to go home .They were all ready for their journey. \\
    \hline
    kG Story & We went to the park. It was a great day. There were so many cool statue. This man had an amazing time. I walked around. \\
    \hline
    MCSM Model & A group of friends decide to spend the day in the park. This is a skateboarder [male], he looks amazing in the park. They went on a skateboard ride and saw this! female looks amazing in her orange tree trunks. At the end of the day, they had a great time. \\
    \hline
    VIST-GPT v1 & I went to the skatepark yesterday. I brought my friend with me. We had a great time there. We brought our skateboards with us. We did a lot of tricks.  \\
    \hline
    VIST-GPT v2 & The two friends decided to spend the day at the skate park. They met up with some other friends and practiced their skate moves. One of the skaters tried to do an ollie but he just jumped a little bit. His friend tried to do an ollie and he nailed it! The friend then tried to do an ollie on the stairs at the skate park. \\
    \hline
\end{tabular}
% \end{center}

    \caption{Comparison of human-written story from the VIST dataset and model-generated stories, illustrating differences in narrative generation approaches.}
    \label{fig:skatepark_sample}
\end{figure*}

\section{Evaluation}
\label{sec:evaluation}

In the evaluation section, we assess the performance of the proposed model using a variety of metrics to measure its effectiveness in visual storytelling. These metrics gauge story quality by comparing generated outputs to human-written narratives, focusing on visual grounding, coherence, and repetition, such as visual grounding, coherence, and repetition. We employ \textit{both reference-based }and reference-free methods for a comprehensive assessment of how closely our model’s outputs align with human narratives.

\subsection{Reference-Based Metrics}
\label{sec:evaluation:challenges}
% The evaluation methods aim to judge how well generated stories align with images they are based on. Here's a breakdown:
% \begin{enumerate}

    % \textbf{Old Reference-Based Pattern-Matching Metrics:} Some of the traditional methods used as evaluation metrics are this task used pattern matching like BLEU, ROUGE, METEOR and CIDEr. As these were the reference based metrics so they used to work to match the reference stories with the generated one by analyzing words and phrases(n-grams). These traditional evaluation metrics rely on n-gram matching have poor correlation with human evaluation score. They are more suitable for tasks like machine translation, where there is a clear target text to generate. In visual storytelling, however, there can be several equally good stories for the same image sequence.
\textbf{Old Reference-Based (Pattern-Matching) Metrics:} Traditional evaluation metrics for this task, such as BLEU, ROUGE, METEOR, and CIDEr, rely on pattern matching with reference-based approaches. These metrics assess the similarity between generated stories and reference stories by analyzing matching words and phrases (n-grams). However, these traditional metrics, which focus on n-gram overlap, often correlate poorly with human evaluation scores. They are more suitable for tasks like machine translation, where a specific target text serves as a clear reference. In visual storytelling, by contrast, there may be multiple equally valid stories for the same sequence of images, making rigid reference-based matching less effective.

\subsection{Reference-free Metrics}

% In our work, we leveraged reference-free metrics specifically designed to evaluate visual storytelling as follows.

% We evaluated our models using Visual Grounding (via GROOViST) \cite{surikuchi2023groovist}, Coherence (via RoViST-C) and Non-Redundancy (via RoViST-NR) \cite{wang2022rovist}. We obtained the average score for the same 900 examples in the test set and compared these results with those of various previously released models. The results are presented in table \ref{tab:metrics} (the higher the score the better). The models we compared against include AREL \cite{wang2018no}, GLACNET \cite{kim2018glac}, KG-Story \cite{hsu2020knowledge}, and MCSM+BART \cite{chen2021commonsense}.

\begin{enumerate}

    \item \textbf{RoViST \cite{wang2022rovist}:} It focuses on three main aspects of generated stories:
    \begin{itemize}
        \item \textbf{Visual Grounding:} How well the story is connected to the visual content of the images.
        \item \textbf{Coherence:} How logically and smoothly the sentences of the story follow each other.
        \item \textbf{Repetition:} How often the same ideas or words are repeated within the story.
    \end{itemize}

    \item \textbf{RoViST-VG (for Visual Grounding):}
        RoViST-VG measures how well a story's sentences are visually grounded in the content of related images. It focuses on aligning nouns in the text with specific regions in the images by projecting both into a shared embedding space using Vision Transformers (ViT) for images and GLoVe embeddings for nouns. The model computes cosine similarity between the noun and image region embeddings. The resulting visual grounding score reflects how closely the story's nouns connect to the visual content in the images, indicating how well the story aligns with the image sequence.

    \item \textbf{RoViST-C (for Coherence):}
        RoViST-C measures how coherent the story is—how well each sentence follows logically from the previous ones.
        It uses a fine-tuned version of the ALBERT model (a type of AI language model) to predict the likelihood that each sentence in the story follows the earlier sentences.
        A story is more coherent if its sentences flow logically from one to the next, and RoViST-C gives a higher score to stories that are more coherent in this way.

    \item \textbf{RoViST-NR (for Non-Redundancy):}
        RoViST-NR measures the degree of non-redundancy in the story, which means how well the story avoids repeating the same information or phrases.
        It calculates the Jaccard Similarity between words and phrases in different parts of the story. If two parts of the story (or phrases within a sentence) have a lot of overlapping words, the story is considered repetitive.
        A story gets a higher non-redundancy score if it has less repetition, meaning that it avoids unnecessary repetition of words or ideas.

    \item \textbf{GROOViST \cite{surikuchi2023groovist}:}
        It is a more advanced metric that focuses specifically on visual grounding, which is about how well the objects and events mentioned in the story correspond to the objects in the images.
        GROOViST checks if the nouns or key objects in the story match the objects shown in the images by using CLIP embeddings (a powerful model that can understand both images and text).

    \item \textbf{Correlation to human judegment:}
    When analyzing the correlation of these metrics with human judgment scores, they show higher correlation compared to previous metrics.

\end{enumerate}

\begin{table*}[htbp] % Use [htbp] for flexible positioning
\centering
\setlength{\tabcolsep}{10pt} % Adjust column separation
\renewcommand{\arraystretch}{1.3} % Adjust row separation
\begin{tabular}{|c|c|c|c|}
    \hline
    \textbf{Model} & \textbf{Visual Grounding Score} & \textbf{Coherence Score} & \textbf{Non-Redundancy Score} \\
    \hline
    AREL & 0.6001 & 0.5692 & 0.8325 \\
    \hline
    GLACNET & 0.5158 & 0.6875 & 0.9506 \\
    \hline
    KG Story & 0.7325 & 0.6493 & \textbf{0.9991} \\
    \hline
    MCSM+BART & 0.8648 & 0.6651 & 0.8999 \\
    \hline
    % GIT Model & \textbf{1.0634} & 0.4463 & 0.6833 \\
    % \hline
    % LLaVa LLaMA 3 & 0.9640 & 0.7026 & 0.9149 \\
    % \hline
    VIST-GPT v1 & 0.9401 & 0.7495 & 0.8821 \\
    \hline
    VIST-GPT v2 & \textbf{0.9962} & \textbf{0.7837} & 0.9301 \\
    \hline
\end{tabular}
\caption{Comparison of models performances on different metrics: Visual Grounding (via GROOViST), Coherence (via RoViST-C), and Non-Redundancy (via RoViST-NR) on the same 900 test set.}
\label{tab:model_performance}
\end{table*}

% \begin{tabular}{|c|c|c|c|c|c|}
%     \hline
%     \textbf{Model} & \textbf{Visual Grounding (GROOVIST)} & \textbf{Coherence (RoVIST-C)} & \textbf{Non-Redundancy (RoVIST-NR)} & \textbf{Intra-Repetition} & \textbf{Avg Story Len} \\
%     \hline
%     AREL (Wang et al., 2018) & 0.584 & 0.577 & 0.833 & 23.5 & 39.1 \\
%     \hline
%     MCSM+BART (Chen et al., 2021) & 0.852 & 0.666 & 0.865 & 2.8 & 56.7 \\
%     \hline
%     TAPM (Yu et al., 2021) & 0.734 & 0.671 & 0.903 & 6.8 & 45.0 \\
%     \hline
%     BLIP2 + GPT-4 (Li et al., 2023) & 0.556 & 0.722 & 0.871 & 1.2 & 175.5 \\
%     \hline
%     LLaVA + GPT-4 (Liu et al., 2023) & 0.653 & 0.759 & 0.810 & 1.4 & 179.2 \\
%     \hline
%     LLaVA w/ SFT & 0.541 & 0.809 & 0.851 & 6.5 & 171.9 \\
%     \hline
%     StoryLLM (Pre-trained Only) & 0.357 & 0.189 & 0.200 & 18.1 & 40.5 \\
%     \hline
%     StoryLLM w/ SFT & 0.578 & 0.772 & 0.856 & 3.4 & 170.7 \\
%     \hline
%     StoryLLM w/ DPO (Ours) & 0.764 & \textbf{0.833} & \textbf{0.905} & \textbf{0.5} & 160.6 \\
%     \hline
%     GLACNET & 0.5158 & 0.6875 & 0.9506 & - & - \\
%     \hline
%     KG Story & 0.7325 & 0.6493 & \textbf{0.9991} & - & - \\
%     \hline
%     VIST-GPT v1 & 0.9401 & 0.7495 & 0.8821 & - & - \\
%     \hline
%     VIST-GPT v2 (Ours) & \textbf{0.9962} & 0.7837 & 0.9301 & - & - \\
%     \hline
% \end{tabular}
% \caption{Comparison of models' performances on different metrics: Visual Grounding (GROOViST), Coherence (RoViST-C), Non-Redundancy (RoViST-NR), Intra-Repetition, and Average Story Length.}
% \label{tab:model_performance}
% \end{table*}
\subsection{Distance between Humans and Models}

\begin{enumerate}

\item \textbf{Combining RoViST and GROOViST}

Combining both, they measure the quality of the generated stories by their closeness to human-written stories across different dimensions. They compute scores for both the model-generated and human-generated stories for each dimension, then measure the human-model distance for each one. These distances are then aggregated to obtain an overall distance score. The lower the overall distance, the closer the generated stories are to the features of human-written stories.

They used the three dimensions proposed by RoViST—visual grounding, coherence, and repetition—as a reference-free metric, but replacing the visual grounding from RoViST by the more advanced GROOVIST.

% \subsection{Distance between Humans and Models}

\item \textbf{Distance between Humans and Models \cite{surikuchi2024not}}

% \label{sec:3_1}

They compute the visual grounding $G$, coherence $C$, and repetition $R$ scores for a given image sequence and the story generated by the model $M$, and do the same for the human story $H$. Then, they calculate the absolute differences between the scores of the human and model-generated stories for each dimension as shown below:

\begin{equation}
\label{eq:3a}
\begin{split}
    \text{d}^{C}_{HM}&=|C_H - C_M|,\\
    \text{d}^{G}_{HM}&=|G_H - G_M|,\\
    \text{d}^{R}_{HM}&=|R_H - R_M|
\end{split}
\end{equation}

Finally, they compute the distance between the model-generated story and the human-written story as the final metric score.
\begin{equation}
\label{eq:3b}
    \text{d}_{HM}=(\text{d}^{C}_{HM} + \text{d}^{G}_{HM} + \text{d}^{R}_{HM}) / 3
\end{equation}

\item \textbf{Results}

% In our work, we leveraged reference-free metrics specifically designed to evaluate visual storytelling, as follows. We evaluated our models using Visual Grounding (via GROOViST) \cite{surikuchi2023groovist}, Coherence (via RoViST-C), and Non-Redundancy (via RoViST-NR) \cite{wang2022rovist}. We obtained the average score for the same 900 examples in the test set and compared these results with those of several previously released models. The results are presented in Table \ref{tab:model_performance} (where a higher score indicates better performance). The models we compared against include AREL \cite{wang2018no}, GLACNET \cite{kim2018glac}, KG-Story \cite{hsu2020knowledge}, and MCSM+BART \cite{chen2021commonsense}.

In our work, we leveraged reference-free metrics specifically designed to evaluate visual storytelling. We assessed our models using GROOVIST \cite{surikuchi2023groovist} for visual grounding, RoViST-C for coherence, and RoViST-NR for non-redundancy \cite{wang2022rovist}. These metrics collectively provide a comprehensive evaluation of narrative quality, ensuring that the generated stories are visually grounded, logically coherent, and free from unnecessary repetition. To ensure a fair comparison, we obtained predictions for the VIST test set from previous models. Since the test set predictions were incomplete for all models, we took the intersection of the available prediction samples, resulting in 900 examples. We then computed the average scores for these 900 examples and compared our results with those of several previously released models, including AREL \cite{wang2018no}, GLACNET \cite{kim2018glac}, KG-Story \cite{hsu2020knowledge}, and MCSM+BART \cite{chen2021commonsense}. The results, presented in Table \ref{tab:model_performance}, show that a higher score indicates better performance."

% we got the predictions of the VIST dataset from previous models, since it was not complete, we got the intersection between the availabe prediction samples, and resulted with 900 examples. we compared all the models with these 900 examples for fair comparison.

As shown in the table \ref{tab:model_performance}, VIST-GPT v2 attained the highest visual grounding and coherence score. Additionally, KG Story achieved the highest score in non-redundancy.

VIST-GPT v2 stands out with excellent scores in all metrics. Its visual grounding score of 0.996 reflects a strong alignment between narratives and visual content, while a coherence score of 0.784 indicates well-structured stories. With a non-redundancy score of 0.930, it generates diverse narratives without repetition. This well-rounded performance positions VIST-GPT v2 as a top choice for effective visual storytelling.

We calculated the Human to Machine Distance $d_{HM}$ for each model, as shown in Figure \ref{fig:dhm} and Table \ref{tab:model_dhm}. VIST-GPT v1, and VIST-GPT v2 exhibit the lowest distances, indicating that they closely align with human expectations in visual storytelling. These models perform well across all three metrics, with VIST-GPT v2 demonstrating the strongest overall performance. This suggests that these models effectively generate coherent narratives that maintain visual relevance without redundancy.

\begin{figure}[htbp]
    \centering
    \includegraphics[width=0.48\textwidth]{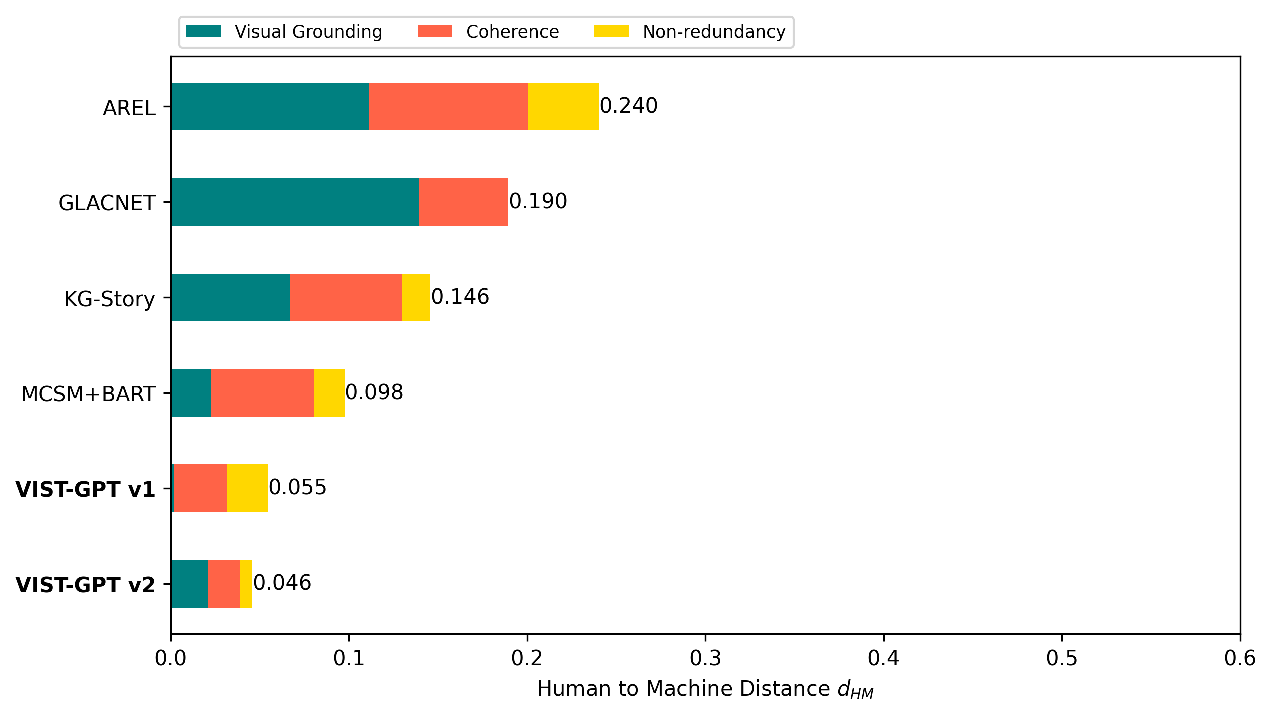}
    \caption{Human to Machine Distance $d_{HM}$ across different models (the less the better)}
    \label{fig:dhm}
\end{figure}

\end{enumerate}

\subsection{UniEval Evaluation}

\begin{table}[htbp] % Use [htbp] for flexible positioning
\centering
\setlength{\tabcolsep}{10pt} % Adjust column separation
\renewcommand{\arraystretch}{1.3} % Adjust row separation
\begin{tabular}{|c|c|c|c|}
    \hline
    \textbf{Model} & \textbf{$d_{HM}$} \\
    \hline
    AREL & 0.2403 \\
    \hline
    GLACNET & 0.1896 \\
    \hline
    KG Story & 0.1457 \\
    \hline
    MCSM+BART & 0.0976 \\
    \hline
    % GIT Model & 0.263057541 \\
    % \hline
    % LLaVa LLaMA 3 & 0.067277809 \\
    % \hline
    VIST-GPT v1 & 0.0546 \\
    \hline
    VIST-GPT v2 & \textbf{0.0459} \\
    \hline
\end{tabular}
\caption{Comparison of $d_{HM}$ for each model (the lower the better), evaluated on the same 900 test set.}
\label{tab:model_dhm}
\end{table}

UniEval \cite{zhong2022towards} serves as a comprehensive evaluation framework that assesses Natural Language Generation (NLG) on multiple dimensions—coherence, consistency, fluency, and relevance—moving beyond traditional metrics like ROUGE and BLEU that fail to capture deeper textual attributes. By reframing evaluation as a binary question-answering task (e.g., “Is this a coherent summary?”), UniEval effectively evaluates NLG outputs with a single model guided by distinct queries. This unified approach, combined with intermediate multitask learning and zero-shot capability, enables UniEval to deliver performance that aligns closely with human evaluations. 

\begin{figure}[htbp] % Use [htbp] for flexible positioning
    \centering
    \includegraphics[width=0.48\textwidth]{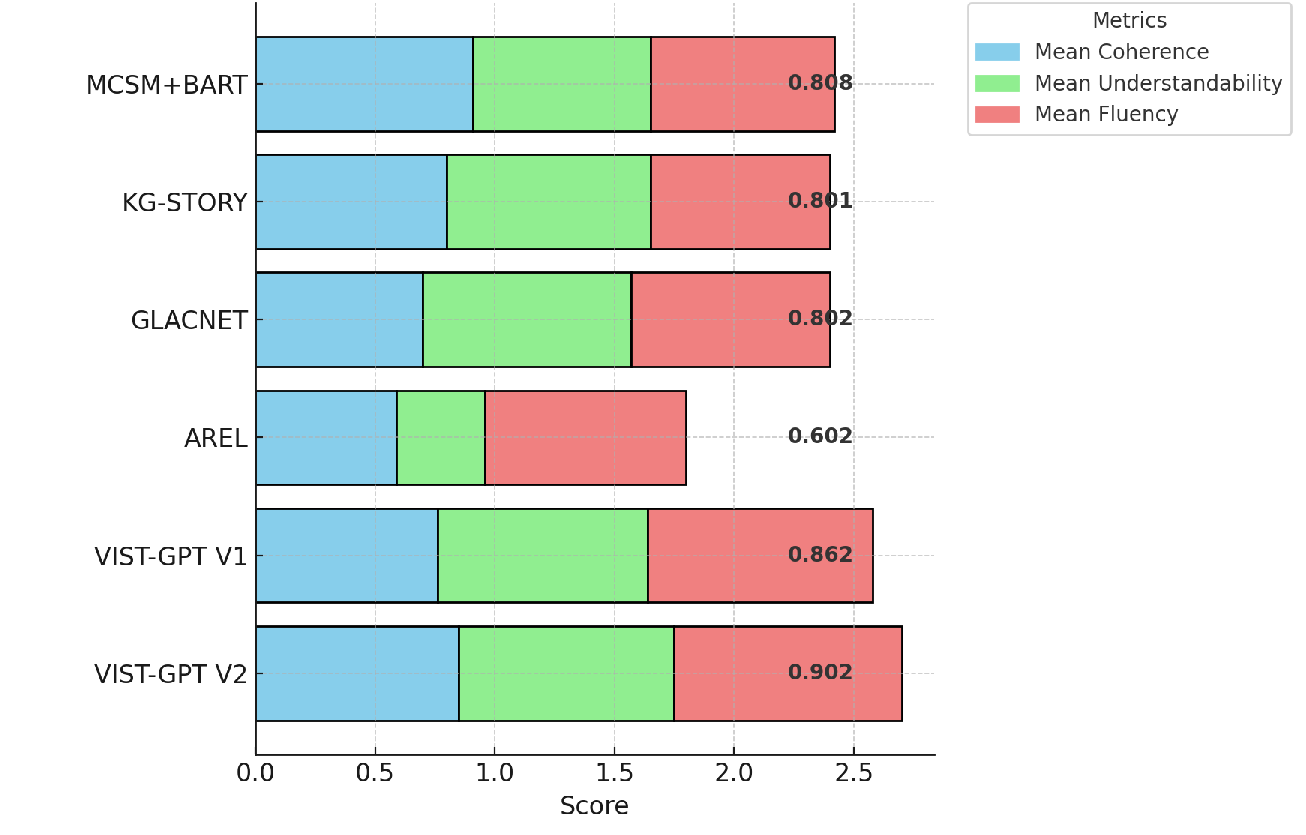}

    \caption{Comparison of coherence, understandability, and fluency across different models.}
    \label{fig:unieval}
\end{figure}

In our paper, we employed UniEval to evaluate various models in the task of visual storytelling, focusing on the coherence, understandability, and fluency of the generated narratives. As shown  in figure \ref{fig:unieval}, our model, VIST-GPT, in two variants(V1 and V2), demonstrates substantial improvement over baseline models in these aspects. 

In terms of \textbf{Mean Coherence}, VIST-GPT V2 achieves a remarkable score of 0.855, reflecting its strong ability to generate logically consistent narratives. While MCSM+BART leads slightly with a score of 0.914, VIST-GPT V2 outperforms other models such as KG-STORY (0.804) and GLACNet (0.704) by a significant margin. This improvement demonstrates the effectiveness of the refinements in our model for generating coherent stories. 

For \textbf{
Mean Understandability}, VIST-GPT V2 achieves the highest score of 0.9, surpassing all baseline models. This result highlights the model's ability to produce narratives that are easy to comprehend, making it particularly effective for visual storytelling tasks. It outperforms GLACNet (0.874), KG-STORY (0.853), and other models, demonstrating its superior interpretability. 

The \textbf{Mean Fluency} metric further underscores the strength of VIST-GPT V2, with a score of 0.950—the highest among all models. This indicates that the language generated by VIST-GPT V2 is not only natural and smooth but also significantly more polished than that of the baseline models, including MCSM+BART (0.771), KG-STORY (0.745), and GLACNet (0.829). The improvements in fluency reflect the success of our model in producing linguistically sophisticated outputs. 

VIST-GPT V2 also shows substantial improvements over its earlier variant, VIST-GPT V1, across all three metrics. This progression highlights the impact of the enhancements made during the model development process. VIST-GPT V2 consistently performs better, showcasing its robustness and reliability as a state-of-the-art visual storytelling model. 

In conclusion, the results demonstrate that VIST-GPT V2 sets a new benchmark for visual storytelling by achieving high scores across coherence, understandability, and fluency. Its ability to generate coherent, interpretable, and linguistically refined narratives makes it a valuable contribution to the field. These advancements position VIST-GPT V2 as a leading model for visual storytelling tasks. 

We evaluated our models and compared their performance to previous models. As shown in Figure \ref{fig:unieval}, we assessed key dimensions such as coherence, understandability, and fluency across various models. These specific metrics were selected because they do not require contextual information for evaluation, allowing for a more direct and consistent comparison of results.

\section{Discussion}
\label{sec:discussion}

VIST-GPT substantially improves visual storytelling, achieving strong performance across key dimensions. Its dual-encoder architecture and adaptive pooling strategies efficiently handle both spatial and temporal aspects of visual input, resulting in coherent and contextually relevant narratives.

The model excels in understanding image sequences, and minimizing information loss while processing visual data. This approach allows for enhanced temporal representation, enabling the generation of more accurate and diverse stories without redundancy.

Despite its impressive performance, the model still encounters challenges in capturing fine-grained details in complex visual inputs. Leveraging more sophisticated large language models (LLMs) or incorporating diverse, domain-specific storytelling datasets could enhance its ability to generate richer, more nuanced narratives. These refinements would likely improve the model's adaptability to a wider range of visual scenarios, ensuring more precise and contextually relevant story generation. In addition, we consider exploring augmentation strategies, like those outlined in \cite{Aboudeshish2025augmentation}, to improve the diversity of the VIST dataset and boost VIST-GPT performance.

% In summary, VIST-GPT’s architectural improvements and training strategies make it a robust model for multimodal storytelling., it outperforms state-of-the-art systems in generating diverse and engaging visual narratives.

In summary, VIST-GPT’s architectural improvements and training strategies make it a robust model for multimodal storytelling. It outperforms state-of-the-art systems in generating diverse and engaging visual narratives.

\section{Conclusion}
\label{sec:conclusion}

% VIST-GPT offers a robust solution for visual storytelling, excelling in grounding, coherence, and non-redundancy. Future enhancements in visual processing and narrative structuring can further elevate its performance, making it an ideal candidate for multimodal generation tasks, achieving state of the art performance.

% VIST-GPT, a model for visual storytelling that excels in visual grounding, coherence, and non-redundancy across sequential images. Through a carefully structured approach combining state-of-the-art computer vision techniques with advanced language generation methods, we created a system that can translate sequences of images into cohesive and engaging narratives.

% The model was trained and evaluated to ensure that it not only recognizes and accurately describes visual elements but also preserves the temporal sequence and emotional depth necessary for effective storytelling. By incorporating unique metrics like Human-to-Machine Distance (dHM) for coherence and groundedness, we achieved a storytelling flow that feels natural and closely aligned with human interpretation.

% This work has demonstrated VIST-GPT’s robustness and versatility for multimodal generation tasks, proving its value as an adaptable tool for applications across media, education, and interactive technologies. Future iterations will aim to enhance the model's ability to capture nuanced emotions and complex storylines, ultimately bringing visual storytelling closer to a true human-like experience.

VIST-GPT offers a robust solution for visual storytelling, excelling in grounding, coherence, and non-redundancy. By combining computer vision techniques with advanced language generation, we created a system capable of translating image sequences into cohesive and engaging narratives.

The model was carefully trained to not only recognize and describe visual elements accurately but also to maintain temporal sequence and emotional depth essential for effective storytelling. By incorporating metrics like Human-to-Machine Distance (dHM) for coherence and groundedness, VIST-GPT achieves a storytelling flow that closely aligns with human interpretation.

Our work demonstrates VIST-GPT’s versatility for multimodal generation tasks, highlighting its potential across media, education, and interactive technologies. Future iterations will focus on enhancing the model's ability to capture nuanced emotions and complex storylines, bringing visual storytelling closer to a human-like experience. 
% \input{sec/appendix}
%\\ \\ \\ \\ \\ \\

{
    \small
    \bibliographystyle{ieeenat_fullname}
    \bibliography{main}
}

%\appendix  % This tells LaTeX to switch to appendix mode
% \input{sec/appendix}  % Title of your appendix section
% Now add content for the appendix here
\twocolumn[
\begin{center}
    {\LARGE \textbf{VIST-GPT: Ushering in the Era of Visual Storytelling with LLMs?}}\\
    \vspace{0.3cm} % Adjust the space between title and subtitle
    {\large \textit{Supplementary Material}} % Subtitle in italic and slightly smaller
    \vspace{1cm} % Space before the next content
\end{center}
]
\setcounter{section}{7} % Sets the section counter to 8
\setcounter{subsection}{0} % Sets the subsection counter to 7 (next subsection will be 8)

\section{Appendix}

\subsection{Qualitative Evaluation }

To assess the effectiveness of VIST-GPT v2 in visual storytelling, we compared its outputs with those of baseline models (AREL, GLACNET, kG Story, MCSM, and VIST-GPT v1), as well as human-generated stories, using two image sets: a skateboarding session at a park (Figure~\ref{fig:skatepark_sample}) and a family boating trip (Figure~\ref{fig:boating_sample}).

\subsubsection{Skatepark Story }
The skateboarding image set (Figure~\ref{fig:skatepark_sample}) shows two friends performing tricks at a park. The human story effectively captures the scene's energy and camaraderie, detailing specific skateboarding tricks and expressing the narrator's excitement. In contrast, baseline models struggled to reflect the activity depicted. For example, AREL and GLACNET generated stories that were misaligned with the context, describing unrelated scenarios like a ski resort or a day at the beach. The MCSM model recognized skateboarding but introduced irrelevant details, while kG Story mentioned a pleasant day at the park without any connection to the skateboarding activity.

VIST-GPT v2, however, closely aligns with the images by describing friends practicing skateboarding moves and attempting specific tricks, such as an “ollie.” The phrase “nailed it!” subtly conveys excitement and accomplishment, adding depth to the narrative without explicitly labeling emotions. This attention to detail and alignment with the visual content demonstrates VIST-GPT v2's improved ability to capture not only the activities but also the social atmosphere of the scene.

\subsubsection{Boating Story }
The family boating trip images (Figure~\ref{fig:boating_sample}) depict a relaxed multi-generational outing on a lake. The human story provides a rich description of the family’s shared experience, including interactions and conversation, accurately capturing the sense of warmth and connection. Baseline models, however, frequently diverged from this context; GLACNET, for instance, referenced a beach outing, while kG Story mentioned unrelated social interactions. MCSM identified family members on a boat but failed to capture the familial warmth depicted in the images.

VIST-GPT v2, by contrast, describes the family outing with specific reference to the grandmother, her daughter, and grandson, effectively conveying a sense of togetherness and family bonding. By capturing the generational relationships and setting in a cohesive narrative, VIST-GPT v2 aligns closely with the visual content, creating a story that reflects the social and familial atmosphere depicted in the images.

\subsubsection{A Day by the Coast}

The human-written story (Figure~\ref{fig:49990}) captures the atmosphere of a family gathering, focusing on drinking, playing video games, and transitioning to outdoor activities like a game of catch. It highlights group dynamics and the natural flow of events, creating an engaging and contextually rich narrative. In contrast, the AREL model provides a generic description of a party, focusing on vague elements like walking to a park without capturing the essence of indoor gaming or outdoor activities depicted in the images. The GLACNET model also fails to align with the visuals, introducing irrelevant details such as dancing and being "very cold outside." While it references games, the lack of specificity and coherence detracts from its relevance to the image set. The kG Story model focuses on a gathering with family and friends but fails to describe the video gaming or outdoor sports activities, making the narrative feel incomplete and disconnected. Similarly, the MCSM model introduces irrelevant elements, such as a "new computer" setup, which is not supported by the images. While it mentions games, it fails to capture the transition to outdoor activities, leading to a less cohesive story.

 VIST-GPT v2 delivers a well-structured and contextually accurate narrative. It describes two friends having fun and transitioning smoothly to outdoor sports, reflecting the flow of events seen in the images. By avoiding irrelevant or hallucinated details and maintaining coherence throughout, VIST-GPT v2 provides the most engaging and visually aligned narrative compared to the baseline models.

\subsubsection{Family Game Night}

As illustrated in Figure \ref{fig:49985}, the human-written story captures the essence of a casual family gathering, emphasizing drinking, gaming, and transitioning to outdoor activities like a game of catch. It provides a vivid and detailed narrative that aligns closely with the sequence of activities depicted in the images, highlighting group dynamics and smooth transitions between events. In comparison, the AREL model describes a generic party, mentioning preparations and walking to a park. However, it fails to include specific activities such as video gaming or outdoor play, leaving the narrative vague and disconnected from the visual content. Similarly, the GLACNET model introduces irrelevant details, such as dancing and cold weather, which are not depicted in the images. While it briefly mentions games, it lacks the specificity and coherence required to capture the atmosphere of the family gathering.

The kG Story model describes a general family gathering with some mention of games but does not include details about the video gaming session or the outdoor activities, resulting in a narrative that feels incomplete and less aligned with the visuals. The MCSM model introduces unrelated elements, such as a "new computer," which are not supported by the images. While it mentions games, it fails to reflect the natural transition from indoor gaming to outdoor play, making the story less cohesive.

VIST-GPT v1 provides a better alignment by mentioning drinking, video gaming, and group interactions. However, it includes unnecessary and irrelevant details, such as "their daughter wouldn’t leave them alone," which are not supported by the images. Moreover, the narrative lacks a logical flow to outdoor activities, making it feel incomplete.

In contrast, VIST-GPT v2 excels by delivering a coherent and contextually accurate narrative. It captures the group having fun drinking and playing video games indoors before transitioning smoothly to outdoor sports, reflecting the sequence of events in the images. VIST-GPT v2 avoids hallucinated or irrelevant details and maintains a logical and engaging narrative flow, making it the most aligned and contextually appropriate story compared to the baseline models.

\subsubsection{Australia Vacation and Harbor Tour}
As shown in figure \ref{fig:50086}, the human-written story captures the atmosphere of a family gathering, focusing on drinking, playing video games, and transitioning to outdoor activities like a game of catch. It highlights group dynamics and the natural flow of events, creating an engaging and contextually rich narrative. In contrast, the AREL model provides a generic description of a party, focusing on vague elements like walking to a park without capturing the essence of indoor gaming or outdoor activities depicted in the images. The GLACNET model also fails to align with the visuals, introducing irrelevant details such as dancing and being "very cold outside." While it references games, the lack of specificity and coherence detracts from its relevance to the image set. The kG Story model focuses on a gathering with family and friends but fails to describe the video gaming or outdoor sports activities, making the narrative feel incomplete and disconnected. Similarly, the MCSM model introduces irrelevant elements, such as a "new computer" setup, which is not supported by the images. While it mentions games, it fails to capture the transition to outdoor activities, leading to a less cohesive story.

VIST-GPT v1 performs better, referencing drinking, playing video games, and group interactions. However, it introduces unnecessary details, such as “the daughter wouldn’t leave them alone,” which are not depicted in the visuals. Furthermore, the narrative lacks a logical transition to outdoor activities, leaving the story incomplete. In contrast, VIST-GPT v2 delivers a well-structured and contextually accurate narrative. It describes the group having fun drinking, playing video games, and transitioning smoothly to outdoor sports, reflecting the flow of events seen in the images. By avoiding irrelevant or hallucinated details and maintaining coherence throughout, VIST-GPT v2 provides the most engaging and visually aligned narrative compared to the baseline models.

\subsection{VIST-GPT v2 Evaluation}
VIST-GPT v2 stands out as the closest model to achieving human-like storytelling, excelling in several aspects:

\textbf{Contextual Relevance:} VIST-GPT v2 aligns closely with the visual content, referencing specific activities such as dining, visiting a museum to see statues, and exploring the area. Unlike baseline models that introduce irrelevant or generic details (e.g., stained glass, anniversaries), VIST-GPT v2 focuses on the core elements of the images, maintaining high fidelity to the visual cues.

\textbf{Narrative Coherence:} The story generated by VIST-GPT v2 flows naturally, with clear transitions between activities. For example, it mentions how "the next morning they decided to go to an art museum," which adds a temporal structure that mirrors human storytelling. This cohesion makes the story engaging and easy to follow, surpassing the disjointed outputs of other models.

\textbf{Social Dynamics:} VIST-GPT v2 effectively captures the interactions and shared experience between the two individuals depicted in the images. The narrative subtly highlights their companionship through shared decisions ("the two friends decided to go out for drinks") and collective exploration. This emphasis on relationships adds emotional depth to the story.

\textbf{Attention to Setting:} The model incorporates the coastal setting and beach exploration, aligning well with the imagery. While it doesn’t explicitly mention every visual element, such as the bike ride, it maintains a consistent connection to the beach and statues, creating a story that feels both immersive and authentic.

\textbf{Engaging Details:} VIST-GPT v2 includes engaging and contextually appropriate details, such as “losing track of time” and “seeing some interesting statues.” These details add richness and texture to the story, enhancing its narrative depth compared to the more superficial or irrelevant descriptions provided by baseline models.

\textbf{Balanced Specificity:} Unlike VIST-GPT v1, which adds unnecessary and overly specific details (e.g., drinks at a restaurant), VIST-GPT v2 strikes a balance between specificity and generality, ensuring that the story remains relevant to the visuals while avoiding superfluous content.

\paragraph{Comparative Advantages Over Baseline Models}
\textbf{AREL and GLACNET:} These models generate highly generic or unrelated narratives, such as describing a "city visit" or focusing on stained glass, which are completely misaligned with the imagery. In contrast, VIST-GPT v2 remains focused on the beach and social context. 

\textbf{kG Story:} While this model captures some elements of the beach and statues, its lack of transitions and relational dynamics makes the story feel disconnected and impersonal. VIST-GPT v2 improves by incorporating both social interactions and a narrative flow. 

\textbf{MCSM:} The hallucination of details like anniversaries and trails detracts from MCSM’s relevance. VIST-GPT v2 avoids such errors by adhering closely to the visual cues.

\textbf{VIST-GPT v1:} Although VIST-GPT v1 captures some aspects of the scene, such as dining and socializing, its output lacks the cohesion, transitions, and emotional depth seen in VIST-GPT v2.

\subsection{Overall Evaluation of VIST-GPT v2 Across Samples}

Based on the evaluations of the five samples provided (Figures~\ref{fig:skatepark_sample} to \ref{fig:50086}), \textbf{VIST-GPT v2} demonstrates significant strengths in visual storytelling, establishing itself as a highly capable and advanced model for generating coherent, contextually rich, and visually grounded narratives. The following highlights the key strengths of VIST-GPT v2 observed across all samples:

\subsubsection{Contextual Relevance and Visual Alignment}
VIST-GPT v2 consistently outperforms baseline models by generating stories that are highly aligned with the visual content. Across all samples:
\begin{itemize}
    \item In the \textbf{skateboarding sample} (Figure~\ref{fig:skatepark_sample}), VIST-GPT v2 captures specific actions like performing an "ollie" and the social dynamics of friends practicing tricks together, focusing on the skatepark environment without introducing irrelevant details.
    \item For the \textbf{boating trip} (Figure~\ref{fig:boating_sample}), the model describes a multi-generational family outing, referencing specific elements like the grandmother, daughter, and grandson. It highlights activities such as exploring the lake and chatting on the boat, offering an accurate and engaging narrative.
    \item In the \textbf{beachside exploration sample} (Figure~\ref{fig:49990}), VIST-GPT v2 captures the sequence of events, from dining at a restaurant to visiting statues, attending a museum, and ending with a bike ride along the coast. These elements closely mirror the visual sequence.
    \item For the \textbf{family game night} (Figure~\ref{fig:49985}), the model balances indoor activities (drinking and video games) with the transition to outdoor play, such as sports, reflecting the depicted scenes.
    \item In the \textbf{Australian harbor vacation} (Figure~\ref{fig:50086}), the model highlights iconic imagery such as the helicopter with an Australian flag, the harbor, and the museum visit, creating a vivid and accurate representation of the visual content.
\end{itemize}

\subsubsection{Narrative Coherence and Flow}
VIST-GPT v2 excels in creating narratives that are not only accurate but also coherent and logically structured:
\begin{itemize}
    \item It effectively handles temporal progressions, such as "after enjoying the harbor, he decided to check out the museums" in the Australian harbor story or "the next morning they decided to go to an art museum" in the beachside exploration sample.
    \item By maintaining a clear and logical flow, VIST-GPT v2 avoids the fragmented or disjointed outputs observed in baseline models like MCSM and kG Story.
\end{itemize}

\subsubsection{Richness of Details}
VIST-GPT v2 provides engaging and contextually appropriate details, enriching the stories:
\begin{itemize}
    \item In the \textbf{skateboarding sample}, it adds vivid descriptions like "nailing the ollie," reflecting the excitement and accomplishments of the activity.
    \item For the \textbf{boating trip}, it mentions the family structure and transitions to exploring the area around the lake, adding depth to the narrative.
    \item The \textbf{beachside exploration story} includes elements like losing track of time, seeing interesting statues, and ending the day with a bike ride, painting a detailed and immersive picture.
    \item In the \textbf{family game night sample}, it captures the mix of indoor and outdoor activities, balancing details about gaming and outdoor sports.
    \item The \textbf{Australian harbor vacation} describes the helicopter with the Australian flag and the awe-inspiring beauty of the harbor, reflecting the grandeur of the scene.
\end{itemize}

\subsubsection{Avoidance of Hallucinations}
Unlike baseline models (e.g., MCSM, GLACNET), which often introduce irrelevant or hallucinated details such as stained glass, "a bird in the roof," or "anniversaries," VIST-GPT v2 remains grounded in the visual content. It avoids fabricating elements that are not present in the images, ensuring high fidelity to the visuals.

\subsubsection{Capturing Social Dynamics}
VIST-GPT v2 effectively portrays social interactions and relationships:
\begin{itemize}
    \item In the \textbf{boating trip}, it emphasizes the multi-generational family aspect, capturing the warmth and togetherness of the outing.
    \item For the \textbf{skateboarding sample}, it highlights the camaraderie of friends practicing tricks together.
    \item The \textbf{family game night} story captures the essence of a family gathering, describing shared activities and interactions.
    \item The \textbf{beachside exploration} narrative reflects companionship through shared exploration and decision-making.
\end{itemize}

\subsubsection{Engagement and Emotional Depth}
The stories generated by VIST-GPT v2 are engaging and convey subtle emotional tones:
\begin{itemize}
    \item It captures the excitement and accomplishment in the \textbf{skateboarding sample}.
    \item In the \textbf{boating trip}, it portrays a sense of relaxation and bonding among family members.
    \item The \textbf{beachside story} conveys a sense of exploration and wonder, while the \textbf{Australian harbor vacation} emphasizes awe and admiration for the sights.
\end{itemize}
\subsubsection{
Attention to Setting:} The model incorporates the coastal setting and beach exploration, aligning well with the imagery. While it doesn’t explicitly mention every visual element, such as the bike ride, figure \ref{fig:50086}it maintains a consistent connection to the beach and statues, creating a story that feels both immersive and authentic.

\subsubsection{Conclusion}
Across all five samples, \textbf{VIST-GPT v2} consistently delivers contextually accurate, coherent, and engaging narratives that align closely with the visual content. Its ability to capture specific details, social dynamics, and emotional undertones sets it apart from baseline models. VIST-GPT v2 not only avoids the pitfalls of hallucination and generic storytelling but also excels in creating vivid, immersive, and human-like narratives, making it a significant advancement in visual storytelling.

\setcounter{figure}{4} % 
\begin{figure*}[h]
    \centering
    \includegraphics[width=1\textwidth]{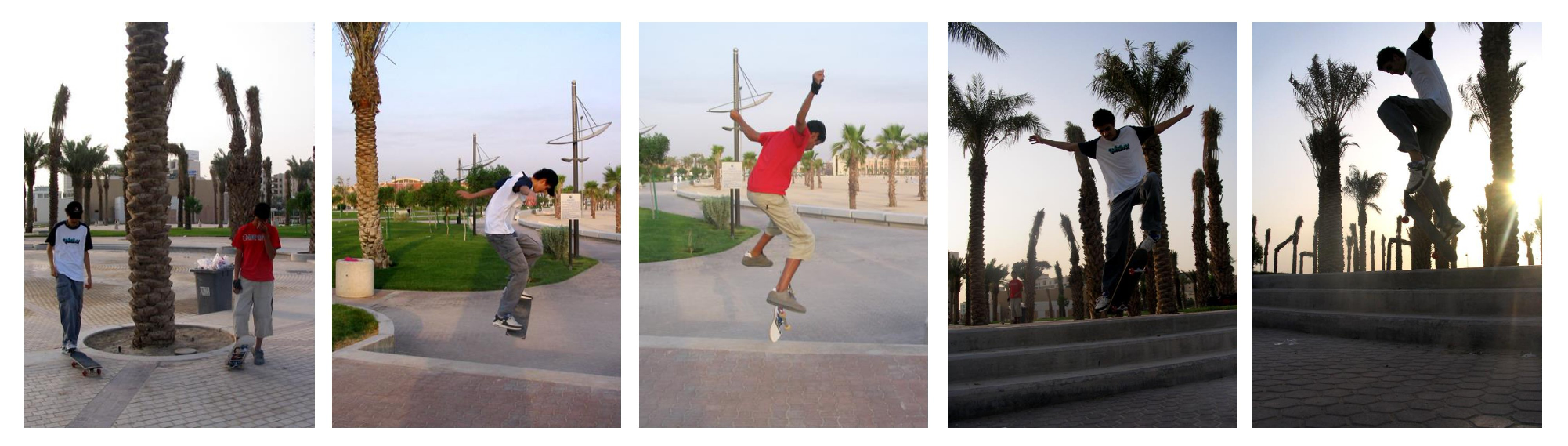}

\begin{tabular}{ l  p{13cm}} % Adjust the width (e.g., 4cm) as needed
    \hline
    Human Story & Hanging out with Abraham, getting some skate time in. Hit an olie of a two step, just warming up. Abraham hit this sweet kickflip! It made for a great photo. Manual up to a three step...into a frontside kickflip! My first time hitting this trick! I was super excited.  \\
    \hline
    AREL Model & I went to the park yesterday. The man is walking down the street to the park. The man in the middle of the race was a bit nervous, but he wasn't sure what he was doing. The man is trying to figure out what to do. This is a picture of a man. \\
    \hline
    GLACNET Model & The man was excited to be at the ski resort. He was very excited for his first day of skiing. He was able to make sure that he wasn't going fast. After a long ride, it was time to go home .They were all ready for their journey. \\
    \hline
    kG Story & We went to the park. It was a great day. There were so many cool statue. This man had an amazing time. I walked around. \\
    \hline
    MCSM Model & A group of friends decide to spend the day in the park. This is a skateboarder [male], he looks amazing in the park. They went on a skateboard ride and saw this! female looks amazing in her orange tree trunks. At the end of the day, they had a great time. \\
    \hline
    VIST-GPT v1 & I went to the skatepark yesterday. I brought my friend with me. We had a great time there. We brought our skateboards with us. We did a lot of tricks.  \\
    \hline
    VIST-GPT v2 & The two friends decided to spend the day at the skate park. They met up with some other friends and practiced their skate moves. One of the skaters tried to do an ollie but he just jumped a little bit. His friend tried to do an ollie and he nailed it! The friend then tried to do an ollie on the stairs at the skate park. \\
    \hline
\end{tabular}
% \end{center}

    \caption{Comparison of human-written and model-generated stories for a "Skatepark Story" scenario from the VIST dataset, illustrating differences in narrative generation approaches.}
    \label{fig:skatepark_sample}
\end{figure*}

\begin{figure*}[h]
    \centering
    \includegraphics[width=1\textwidth]{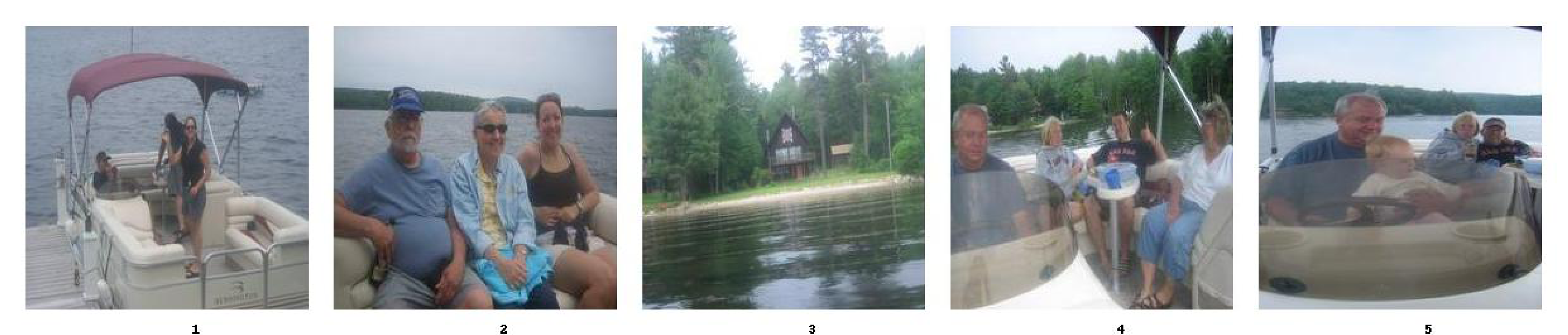}

% \begin{center}
% \begin{tabular}{l c r}
\begin{tabular}{ l  p{13cm}} % Adjust the width (e.g., 4cm) as needed
    \hline
    Human Story & We took the boat out for a ride on the lake. We all piled on for a great conversation and had a few drinks. The captain of this ship is so young, just a baby really. We went by a cabin on the lake, maybe we'll rent that for our next vacation. Everybody had a great time and can't wait to do it again.  \\
    \hline
    AREL Model & The boat went on a boat ride on the lake. The family took a picture together. Male and his friends decided to go on a boat ride. The view from the top of the lake was beautiful. We had a great time and had a great time. \\
    \hline
    GLACNET Model & The beach was a lot of fun. There were many people there. We had a great time. It was a very nice day. We spent a lot of time at the beach. \\
    \hline
    KG Story & I love to hang out with my friends. It is a beautiful day. There are so many people in the house. Some of them were very tall. We got back home. \\
    \hline
    MCSM Model & Male and male were on a boat trip and were happy. Male was sitting on the boat and just enjoying the view. Dad and male were there and were surprised to see that mom and dad had already left. While they were at the lake house they stopped to park and ride for a bit. Male and male had a great time on the boat trip. \\
    \hline
    VIST-GPT v1 & We took the boat out on the lake today. We had a great time. The weather was a bit overcast. But it was very warmish out on the water. We had a great time chatting on the boat.  \\
    \hline
    VIST-GPT v2 & The family went on a boating trip. The family consisted of a grandmother, her daughter, and her grandson. The grandson's dad also came along for the ride. The family had a great time together. They took a while to explore the area around the lake. \\
    \hline
\end{tabular}
% \end{center}

    \caption{Comparison of human-written and model-generated stories for a "Boating Story" scenario from the VIST dataset, illustrating differences in narrative generation approaches.}
    \label{fig:boating_sample}
\end{figure*}

\begin{figure*}[h]
    \centering
    \label{fig:49990}\includegraphics[width=1\textwidth]{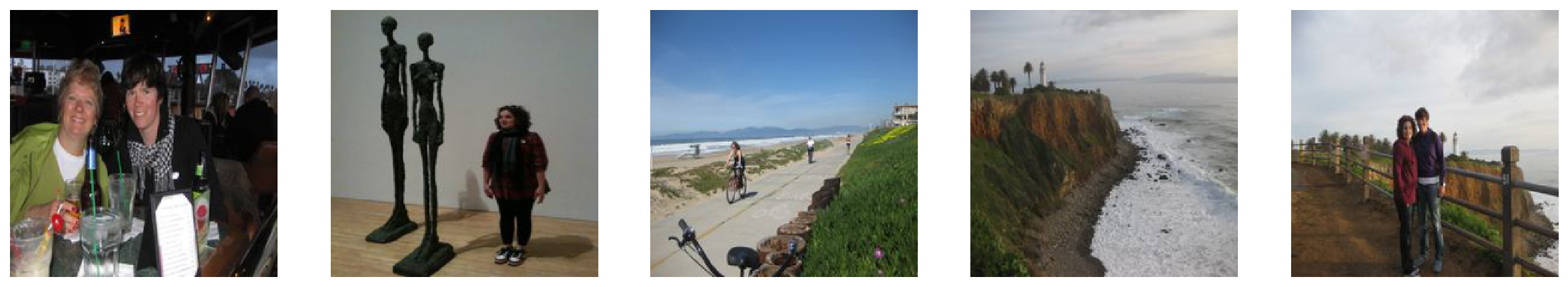}

% \begin{center}
% \begin{tabular}{l c r}
\begin{tabular}{ l  p{13cm}} % Adjust the width (e.g., 4cm) as needed
    \hline
    Human Story & Two women sit at a restaurant table and eat their dinner. They walk through the restaurant and see statues standing in a corner. The beach is next and they take a bike ride. The coast of the beach has gorgeous mountains and trees. The two women pose by the ocean for a final picture. \\
    \hline
    AREL Model & We had a great time at the beach. We saw many interesting statues. We rode on the beach. The view from the top of the beach was beautiful. We had a great time at the beach. \\
    \hline
    GLACNET Model & We went to the city today. There were many interesting statues. Some of them were very nice. They were all very beautiful. I had a great time. \\
    \hline
    KG Story & We went to the beach last night. There were many beautiful statues. I took a picture of my favorite bar. Then walked around and saw some amazing stained bridge. \\
    \hline
    MCSM Model & We went to the beach for male and female's anniversary . Male and female made a statue out of all the love made statues. They went to the beach and found this beautiful view. Along the path, we found a trail that led along the beach. Male and female posed for a picture together. \\
    \hline
    VIST-GPT v1 & The two friends decided to have a day out together. They went to a restaurant and ordered some drinks. They were very happy to be able to spend time together. They had a lot of fun at the restaurant. Afterwards, they went to a museum to see some of the exhibits. \\
    \hline
    VIST-GPT v2 & The two friends decided to go out for drinks. They had a few too many and lost track of time. The next morning they decided to go to an art museum. They saw some interesting statues. After the museum they went for a bike ride along the beach. \\
    \hline
\end{tabular}
% \end{center}

    \caption{Comparison of human-written and model-generated stories for a "A Day by the Coast" scenario from the VIST dataset, illustrating differences in narrative generation approaches.}
    \label{fig:49990}
\end{figure*}

\begin{figure*}[h]
    \centering
    \includegraphics[width=1\textwidth]{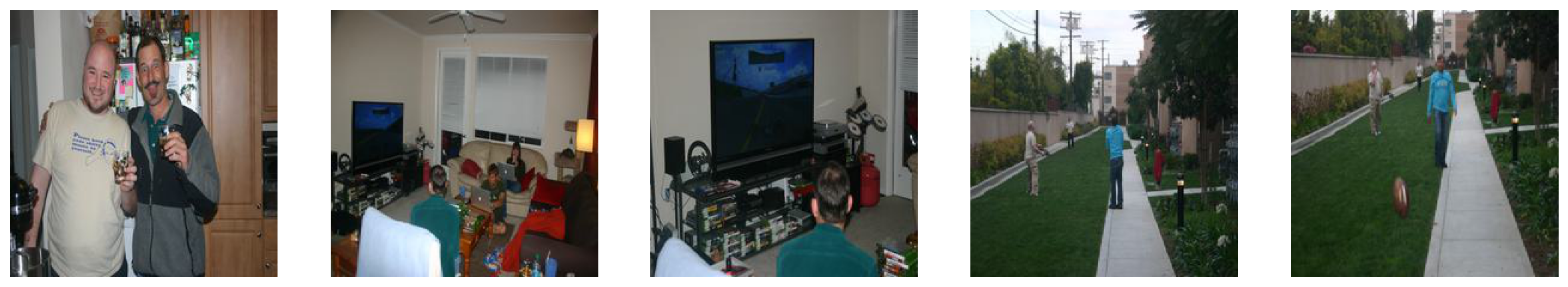}

% \begin{center}
% \begin{tabular}{l c r}
\begin{tabular}{ l  p{13cm}} % Adjust the width (e.g., 4cm) as needed
    \hline
    Human Story & When the guys get together, they like to drink. Everyone sits around the family room playing their own games. Dad is really involved in this racing game. After sitting inside for a while, we decided to go out to get some fresh air. We played a good old fashioned game of catch. \\
    \hline
    AREL Model & We had a great time at the party. We had a lot of fun at the party. The man is getting ready for the party. We walked down the street to the park. He was so excited to be there. \\
    \hline
    GLACNET Model & I had a great time at the party yesterday. There were a lot of people there. We played some games. Afterward we went outside. It was a lot of fun. \\
    \hline
    KG Story & I had a great time at the living room. My family and friends were there. We all gathered around to watch. It was very cold outside. Some of us danced. \\
    \hline
    MCSM Model & Everyone got together for a family get together. Everyone was set up for some games. My uncle got his new computer and was getting ready to play. There were some people sitting down to play some good games. We had a good night and we can't wait to go home. \\
    \hline
    VIST-GPT v1 & The guys wanted to go out for a night on the town. But their daughter wouldn't leave them alone. So they enlisted the help of their other daughter. But their other Daughter wouldn't leave them alone either. So they just gave up and let the kids watch TV while they had a night out. \\
    \hline
    VIST-GPT v2 & We had a party at the apartment. Everyone was having a great time drinking. We watched a few movies on the TV. We also played a few video games. Afterwards we went outside to play some sports. \\
    \hline
\end{tabular}
% \end{center}

    \caption{Comparison of human-written and model-generated stories for a "Family Game Night" scenario from the VIST dataset, illustrating differences in narrative generation approaches.}
    \label{fig:49985}
\end{figure*}

\begin{figure*}[h]
    \centering
    \includegraphics[width=1\textwidth]{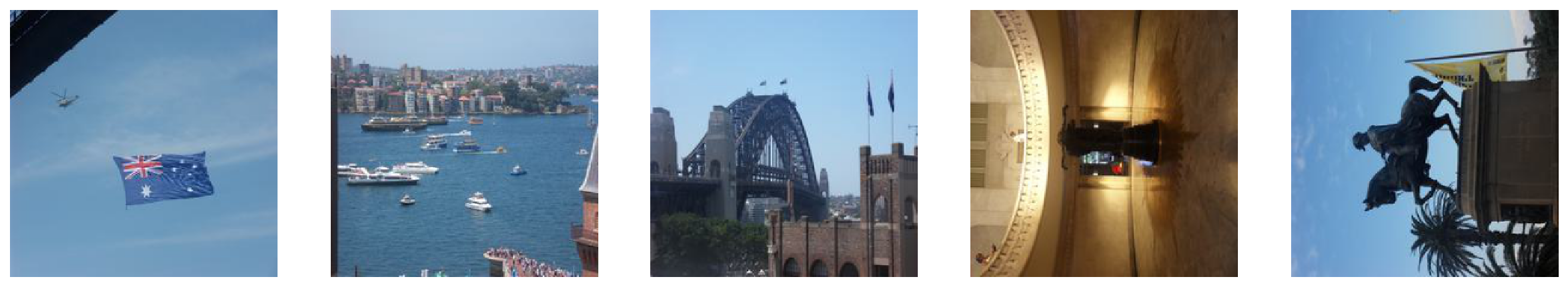}

% \begin{center}
% \begin{tabular}{l c r}
\begin{tabular}{ l  p{13cm}} % Adjust the width (e.g., 4cm) as needed
    \hline
    Human Story & Many people visit Britain, it is a beautiful place to go. Some people love coming in by boat it makes the traveling very nice. Even the bridges are so breathtaking. The view is beyond amazing. Going into museums you can see wonderful works of art. The outside has many statues honoring people. This one for instance, with a man on his horse. \\
    \hline
    AREL Model & we went on a trip to location. The view from the top of the lake was beautiful. This is a picture of a bridge. The inside of the building was very impressive. We had a great time there. \\
    \hline
    GLACNET Model & The plane was very tall. We went to see the boats. There were many different buildings. We had a great time. It was a lot of fun. \\
    \hline
    KG Story & We went to a show. There were many boats. The inside was very cool and saw an old building. It's amazing. \\
    \hline
    MCSM Model & Flying over location location is a great way to celebrate freedom. We can see the flags flying above the water. We can see the boats and people from all around the water. We can even see the buildings from the water. It 's so cool to see all the different kinds of buildings on the water. You can even see the old building with the bird in the roof. \\
    \hline
    VIST-GPT v1 & The helicopter approached the bridge. It was a beautiful day for flying. The clear blue sky was perfect for the helicopter to fly in. The helicopter pilot was very skilled at flying. He flew over the city and the bay. \\
    \hline
    VIST-GPT v2 & The man was on vacation in Australia and was excited to see the sights. He saw a helicopter with an Australian flag attached flying over the bridge his hotel was located on. He decided to go down to the water to watch the boats pass by. He was in awe of how beautiful the harbor was. After he was done enjoying the harbor, he decided to check out the museums in the area. \\
    \hline
\end{tabular}
% \end{center}

    \caption{Comparison of human-written and model-generated stories for a Australia Vacation and Harbor Tour" scenario from the VIST dataset, illustrating differences in narrative generation approaches.}
    \label{fig:50086}
\end{figure*}


\end{document}